\documentclass{article}

\usepackage{microtype}
\usepackage{graphicx}
\usepackage{subfigure}
\usepackage{booktabs} % for professional tables

\usepackage{amsmath}
\usepackage{microtype}

\usepackage[acronym]{glossaries}
\newacronym{FDGATII}{FDGATII}{Fast Dynamic Graph Attention with Initial residual and Identity mapping}
\newacronym{GNN}{GNN}{Graph Neural Networks}
\newacronym{GAT}{GAT}{Graph Attention Networks}
\newacronym{GCN}{GCN}{Graph Convolutional Networks}
\newacronym{CNN}{CNN}{Convolutional Neural Networks}
\newacronym{GCNII}{GCNII}{Graph Convolutional Network via Initial residual
and Identity mapping}

\usepackage{hyperref}

\usepackage[accepted]{icml2021}

\icmltitlerunning{ \textcolor{red}{preprint 2.4} \acrfull{FDGATII}}

\begin{document}

\twocolumn[
\icmltitle{FDGATII : Fast Dynamic Graph Attention with \\Initial Residual and Identity Mapping}

\icmlsetsymbol{equal}{*}

\begin{icmlauthorlist}
\icmlauthor{Gayan K. Kulatilleke}{uq}
\icmlauthor{Marius Portmann}{uq}
\icmlauthor{Ryan Ko}{uq}
\icmlauthor{Shekhar S. Chandra}{uq}
\end{icmlauthorlist}

\icmlaffiliation{uq}{School of Information Technology and Electrical Engineering, Faculty of Engineering, Architecture and Information Technology, University of Queensland, Queensland, Australia}

\icmlcorrespondingauthor{Gayan K. Kulatilleke}{g.kulatilleke@uqconnect.edu.au}
\icmlcorrespondingauthor{Shekhar S. Chandra}{shekhar.chandra@uq.edu.au}

% You may provide any keywords that you
% find helpful for describing your paper; these are used to populate
% the "keywords" metadata in the PDF but will not be shown in the document
\icmlkeywords{Machine Learning, ICML, attention, node classification, heterophilic}

\vskip 0.3in
]

% this must go after the closing bracket ] following \twocolumn[ ...

% This command actually creates the footnote in the first column
% listing the affiliations and the copyright notice.
% The command takes one argument, which is text to display at the start of the footnote.
% The \icmlEqualContribution command is standard text for equal contribution.
% Remove it (just {}) if you do not need this facility.

%\printAffiliationsAndNotice{}  % leave blank if no need to mention equal contribution
\printAffiliationsAndNotice{\icmlEqualContribution} % otherwise use the standard text.

\begin{abstract}
While \acrlong{GNN} have gained popularity in multiple domains, graph-structured input remains a major challenge due to (a) over-smoothing,  (b) noisy neighbours (heterophily), and (c) the suspended animation problem. To address \textit{all these problems simultaneously}, we propose a novel graph neural network FDGATII, inspired by attention mechanism’s ability to focus on selective information supplemented with two feature preserving mechanisms. FDGATII combines Initial Residuals and Identity Mapping with the more expressive dynamic self-attention to handle noise prevalent from the neighbourhoods in heterophilic data sets. By using sparse dynamic attention, FDGATII is inherently parallelizable in design, whist efficient in operation; thus theoretically able to scale to arbitrary graphs with ease. Our approach has been extensively evaluated on 7 datasets. We show that FDGATII outperforms GAT and GCN based benchmarks in accuracy and performance on fully supervised tasks, obtaining state-of-the-art results on Chameleon and Cornell datasets with zero domain-specific graph pre-processing, and demonstrate its versatility and fairness. 
%Code is available at: XXX
\end{abstract}
 
\section{Introduction} \label{Introduction}
Recently, research on graphs has been receiving more and more attention due to the great expressive power and pervasiveness of graph structured data \cite{zhu2020beyond}. Many interesting irregular domain tasks such as 3D meshes, social networks, telecommunication networks and biological networks  involve data that are not representable in grid-like structures \cite{velivckovic2018graph}. As a unique non-euclidean data structure for machine learning \cite{zhu2020beyond}, graphs can be used to represent a diverse set of feature rich domains, from social and dark web forums \cite{samtani2017exploring} to cryptocurrency blockchains. \citet{kipf2016semi} predicted Facebook friend suggestions; \citet{chen2018measurement} analysed large social attribute sets for node classification; and \citet{samtani2017exploring} derived insights from social network structures and attribute sets to identify key players and malware exploits in dark web forums.

A \acrfull{GNN} performs neighbourhood structure aggregation along with node feature transformation to map each node to an embedding vector \cite{liu2020non}. GNN variants mostly differ in how each node aggregates the representations of its neighbours and combines them with its own representation \cite{brody2021attentive}.\acrfull{GCN} \cite{kipf2016semi} generalize \acrfull{CNN} \cite{lecun1995convolutional} to graph-structured data. \acrfull{GAT} \cite{velivckovic2018graph} uses attention. GraphSage \cite{hamilton2017inductive} applies max pooling. The basic approach is to learn an aggregation of neighbour features into a low dimensional vector \cite{zhou2020graph} for downstream tasks such as node classification, clustering, and link prediction \cite{perozzi2014deepwalk,hamilton2017inductive}. 

However, graph-structured inputs are one of the major challenges of machine learning \cite{bronstein2017geometric, hamilton2017inductive, battaglia2021relational} due to (a) the over-smoothing problem which limits the depth and receptive field of models \cite{chen2020simple}, (b) noise or heterophily in features and class label distribution (nodes with similar features but different in labels) \cite{alon2020bottleneck}, which necessitates attending to features without any prior information during the aggregation process, and (c)  the suspended animation problem \cite{wu2019simplifying}.

As most graphs require the interaction between nodes that are not directly connected this is achieved by stacking multiple GNN layers \cite{alon2020bottleneck}. In practice though, GCNs were observed not to benefit from more than few layers due to over-smoothing: node representations become indistinguishable when the number of layers increases \cite{wu2019simplifying}. As over-smoothing was mostly demonstrated in short-range tasks \cite{chen2020simple}, models such as \acrfull{GCNII} \cite{chen2020simple} were able to extend GCN with extra initial residual representations and identity information to achieve better results while still performing local aggregation.

However, due to unfocused uniform aggregation on the neighbourhood, most of these models, including GCNII, are more suitable only for homophily datasets where nodes linked to each other are more likely to belong in the same class, i.e. the neighbourhoods with very low noise, from an aggregation perspective.  In practice, real-world graphs are also often noisy with connections between unrelated nodes resulting in poor performance in current GNNs. Also, GNNs in general are not able to handle tasks that depend on long-range information due to over-squashing: information from the exponentially growing receptive field being compressed into fixed-length node vectors \cite{alon2020bottleneck} due to its unfocused aggregation mechanism. As a result, whilst GCNII achieves state-of-the-art performance on homophilic datasets such as Cora, accuracy in heterophilic datasets such as Texas and Wisconsin is relatively poor (Table~\ref{table_acc}) \cite{zhu2020beyond}.

Many popular GNN models implicitly assume homophily producing results that may be biased, unfair or erroneous \cite{maurya2021improving}. This can result in the so-called ‘filter bubble’ phenomenon in a recommendation system (reinforcing existing beliefs/views, and downplaying the opposite ones), or make minority groups less visible in social networks \cite{zhu2020beyond}.

On the other hand, \citet{vaswani2017attention} showed that self-attention is sufficient for achieving  state-of-the-art performance on machine translation tasks. GAT \cite{velivckovic2018graph} generalized the attention mechanism for graphs using attention-based neighbourhood aggregation. Importantly, GAT improves on the simple averaging or max pooling of neighbours \cite{kipf2016semi, hamilton2017inductive}, by allowing every node to now compute a weighted average of its neighbours \cite{brody2021attentive}, which is a form of selective aggregation. According to \citet{knyazev2019understanding}, the generalization ability of attention mechanism helps GNNs generalize to larger and noisy graphs. By determining individual attention on each node neighbour, GAT is able to ignore the irrelevant neighbours and focus on the relevant neighbours \cite{alon2020bottleneck}. A refinement, GATv2 \cite{brody2021attentive}, uses a more expressive variant termed dynamic attention where the ranking of attended nodes is better conditioned on the query node by replacing the supposedly monotonic GAT attention function with a universal approximator attention function that is strictly more expressive. However, GAT or GATv2 \textit{alone, in its current form} is not able to handle heterophilic data due to the still present essentially local aggregation operation \cite{liu2020non}.

Thus, it remains an open problem to design efficient GNN models that effectively prevents over-smoothing, suspended animation and noise. As observed by \citet{chen2020simple} it is even unclear whether the network depth is a resource or a burden when designing new GNNs. In this work we proposed a light weight, efficient and parallelizable model based on self attention that addresses these challenges simultaneously. 

Our \acrlong{FDGATII}  (\textbf{\acrshort{FDGATII}}), is an efficient shallow dynamic attention based model that overcomes over smoothing and noisy neighbours simultaneously by combining self attention integrated with two feature preserving mechanisms. At each layer, an initial residual constructs a ‘skip connection’ from the input layer, while ‘identity mapping’ preserves the node's identity via by adding an identity matrix to the weight matrix. Mainly, as demonstrated, nodes are aggregated based on individually focused sparse attention which is able to disregard noise and generalize well to homophilic and heterophilic datasets. FDGATII achieves comparable or state-of-the-art results on various full-supervised tasks and over an order of magnitude training and inference time efficiency with significantly low resources. In our work, we do not perform exhaustive hyper parameter tuning as \citet{alon2020bottleneck} shows that prior work with extensively tuned GNNs to real-world datasets suffer from over-squashing.

The contribution of this work can be summarized as follows:
\begin{itemize}
\item We introduce a novel GNN, namely FDGATII, which can help address the multiple challenges in in real world datasets effectively. FDGATII introduces an enhanced attention mechanism to handle noise and generalize to both homophilic and heterophilic (assortative and disassortative) datasets. 
\item We also show that FDGATII is robust to noise while able to solve the main graph challenges : heterophily, over-smoothing and suspended animation  \textit{simultaneously}. 
\item FDGATII is efficient due to using shallow self-attention. Its sparse implementation is also parallelizable across node neighbour pairs and can be split between multiple GPUs; (further, no eigen decompositions or similar costly matrix operations are required.)
\item As FDGATII is based on GAT, it is also directly applicable to inductive learning problems, including tasks where the model must generalize to completely unseen graphs; and finally
\item We test the effectiveness on a wide range of graph benchmark datasets and compare against both classic and state-of-the-art GNN models to demonstrate that FDGATII consumes over a magnitude less computational resources while maintaining or exceeding state-of-the-art GAT and GCN models; and 
\item By not assuming homophily, FDGATII minimizes its potential negative effects, i.e.: bias and unfairness, whist obtaining state-of-the-art results on Chameleon and Cornell benchmark data sets.
\end{itemize}

\section{Preliminaries and Related Work} \label{Preliminaries}
\subsection{Notations} \label{Notations}
We follow the same notations as \citet{chen2020simple} where $G = (V,E)$ is a simple connected undirected graph with $n$ nodes (denoted $\{1,\ldots ,n\}$) and $m$ edges while $\bar{G} = (V,\bar{E})$ is its self-looped graph. $d_i$ and $d_i+1$ is the degree of node $i$ in $G$ and $\bar{G}$ respectively. Given $A$ as the adjacency matrix and $D$ the degree matrix of $G$, the adjacency matrix and degree matrix of $\bar{G}$ is $\bar{A} = A+I$ and $\bar{D} = D+I$ due to the presence of the self-loop. Every $v \in V$ is represented by a $d$-dimensional feature vector $X_v$ where $X \in R^{n \times d}$ is the  feature matrix. The symmetric positive semi definite \textit{normalized graph Laplacian matrix} is given by $L = I_{n}- D^{-1/2}AD^{-1/2}$ whoes eigen decomposition is $U\Lambda U^T$. $\Lambda$ is the diagonal eigenvalue matrix of $L$. $U \in R^{n \times n}$ is the unitary eigenvector matrix of $L$.

Given signal $x$ and filter $g_{\gamma }(\Lambda) = diag(\gamma)$ the graph convolution operation is $g_\gamma (L)\ast x = U g_\gamma(\Lambda)U^Tx$ where $\gamma \in R^n$ is the vector of spectral filter coefficients.

\subsection{Homophily vs Heterophily} \label{Homophily}
Node classification problem relies on the graph structure and features of the nodes to identify the labels of the node. Under homophily, nodes are assumed to have neighbours with similar features and labels \cite{liu2020non}, thus the cumulative aggregation of node’s self-features with it's neighbours reinforces the signal corresponding to the label and helps to improve accuracy of the predictions. However, in case of heterophily, nodes are assumed to have dissimilar features and labels, thus the cumulative aggregation will reduce the signal and add more noise causing the neural network to learn poorly resulting in poor performance \cite{maurya2021improving, zhu2020beyond}. 

Since many existing GNNs assume strong homophily, they fail to generalize to networks with heterophily. Ego- and neighbour-embedding separation, higher-order neighbourhoods and combination of intermediate representations can help improve the performance of GNN models in heterophily settings \cite{zhu2020beyond}. 

Homophily  denotes the fraction of edges which connects two nodes of the same label \cite{liu2020non}. A higher value (closer to 1) indicates strong homophily, while a lower value (closer to 0) indicates strong heterophily in the dataset.

\subsection{Vanilla GCN} \label{GCN}
\citet{hammond2011wavelets} suggested approximating $g_\gamma(\Lambda)$ by a truncated expansion in terms of  K\textsuperscript{th} order Chebyshev polynomial where $\theta \in\mathbf{R}^{K+1}$ corresponds to a vector of polynomial coefficients. 
\begin{equation}
\mathbf{U}g_{\theta}\left(\Lambda\right)\mathbf{U}^{T}\mathbf{x}\approx
\mathbf{U}\left(\sum_{l = 0}^{K}\theta_l\mathbf{\Lambda }^l\right)\mathbf{U}^T\mathbf{x}
=\left(\sum_{l = 0}^{K}\theta_l\mathbf{L}^l\right)\mathbf{x}
\end{equation}

The vanilla GCN \cite{kipf2016semi} simplifies the convolution operation by setting $K=1,\theta_0=2\theta$ and $\theta_1=-\theta$ to derive the convolution operation $g_{\theta}\ast x =\theta(I+D^{-1/2}AD^{-1/2})x$. GCN also applies the renormalization trick, i.e. use the normalized self-looped adjacency matrix $\bar{P} = \bar{D}^{-1/2}\bar{A}\bar{D}^{-1/2}=(D+I_n)^{-1/2}(A+I_n)(D+I_n)^{-1/2}$. Each convolutional layer (Equation~\ref{eq-GCN}) contains a nonlinear
activation function $\sigma$, typically ReLU.

\begin{equation} 
    \mathbf{H}^{l+1} = \sigma\left(\bar{\mathbf{P}}\mathbf{H}^l\mathbf{W}^l\right) 
    \label{eq-GCN}
\end{equation} 

However, since node embeddings are aggregated recursively from the neighbour embeddings layer by layer, the embedding in the final layer requires all embeddings of upper layers, resulting in high memory cost. Also, GCN gradient update in the full-batch training scheme requires storing all intermediate embeddings, which makes the training unable to extend to large graphs. 

Unfortunately, in all above spectral approaches, the learned filters depend on the Laplacian eigen basis, which depends on the entire graph structure. As a result, a model trained on a specific structure cannot be directly applied to a graph with a different structure \cite{velivckovic2018graph}.

\subsection{GCNII} \label{GCNII}
GCNII \cite{chen2020simple} extends GCN to a deep model by enabling GCN to express a $K$ order polynomial filter of arbitrary coefficients with two simple techniques: initial residual connection and identity mapping. Formally, we define the $l$-th layer of GCNII as:
\begin{equation}
    \mathbf{H}^{l+1}=\sigma\left(\left(\left(1-\alpha_l\right)\bar{\mathbf{P}}\mathbf{H}^l+\alpha_l\mathbf{H}^{0}\right)\left(\left(1-\beta_l\right)\mathbf{I}_{n} +\beta_l\mathbf{W}^l\right)\right)
    \label{eq-GCNII}
\end{equation} 
where $\alpha_l$ and $\beta_l$ are hyperparameters. 

In summary, GCNII 1) combines the smoothed representation $\mathbf{P}\mathbf{H}^l$ with an initial residual connection to the first layer $\mathbf{H}^{(0)}$; and 2) adds an identity mapping $\mathbf{I}_{n}$ to the $l$-th weight matrix $\mathbf{W}^l$. \citet{chen2020simple} shows that the alternative to the skip connection in ResNet \cite{he2016identity} the residual connection \cite{kipf2016semi}, is only able to partially relieve the over-smoothing problem. By using a connection to the initial representation $\mathbf{H}^0$, GCNII ensures that the final representation of each node retains at least a $\alpha_l$ fraction from the input layer. 

Further GCNII builds upon \citet{hardt2016identity} who showed that identity mapping of the form $\mathbf{H}^{l+1}=\mathbf{H}^l(\mathbf{W}^l+\mathbf{I}_{n})$ satisfies the following properties: 1) the optimal weight matrices $\mathbf{W}^l$ have small norms; 2) the only critical point is the global minimum. The first property allows us to put strong regularization on $\mathbf{W}^l$ to avoid over-fitting, while the later is desirable in semi-supervised tasks where training data is limited. 

\citet{oono2019graph} theoretically proved that a $K$-layer GCN's convergence rate depends on $s^K$, where $s$ is the maximum singular value of the weight matrices $\mathbf{W}^l,l = 0,\ldots ,K-1$. GCNII replaces $\mathbf{W}^l$ with $(1-\beta_l)\mathbf{I}_{n}+\beta_l\mathbf{W}^l$ with regularization on $\mathbf{W}^l$, resulting in singular values of $(1-\beta_l)\mathbf{I}_{n}+\beta_l\mathbf{W}^l$ closer to 1, which implies that $s^{K}$ is large, and the information loss is relieved. 

However, as GCNII combines neighbour embeddings by uniformly averaging, its heterophilic performance is relatively poor. Alternatively, a selective aggregation over the neighbourhood allows focusing on relevant nodes \cite{zhu2020beyond}.

\subsection{Attention Mechanism}
\begin{figure}[h]
    \centering \includegraphics{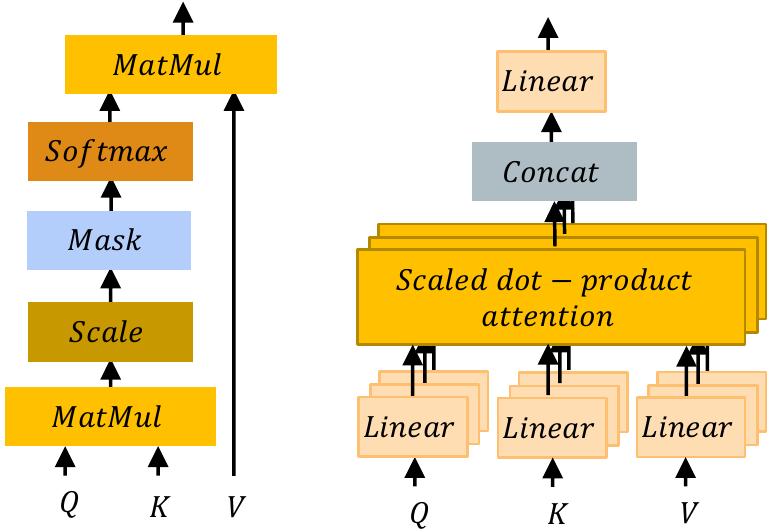} 
    \caption{The \citet{vaswani2017attention} 'Scaled Dot-Product Attention' (left). Multi-Head Attention consists of several attention layers running in parallel (right).}
    \label{fig_attention}
\end{figure}

The attention mechanism \cite{vaswani2017attention} has been widely used in GNNs \cite{chen2020simple,brody2021attentive, velivckovic2018graph}. Attention essentially maps a query $Q$ and a set of key-value pairs $K$, $V$ to an output, where the query, keys, values, and output are all vectors (Figure~\ref{fig_attention}).

\subsection{GAT}
In contrast to GCN, which weighs all neighbours $j\in\mathcal{N}_i$ with equal importance GAT \cite{velivckovic2018graph} computes a learned weighted average of the representations of $\mathcal{N}_i$ using attention. Compared to GCN, attention-based operators assign different weights for neighbours, and can alleviate noises and achieve better results \cite{zhou2020graph} while being more robust \cite{alon2020bottleneck}. 

Specifically, a scoring function $e:R^{d}\times R^{d}\rightarrow R$ computes a score for every edge $ (j,i)$, which indicates the importance of the features of the neighbour $j$ to the node $i$: 
\begin{equation} 
H^{l+1} = \sigma \mathbf{A}\mathbf{h}^l\mathbf{W}^l 
\end{equation} 

\begin{equation} 
H^{l+1} = \sigma \sum_{\mathbf{j}\in \mathcal{N}_{\mathbf{i}}}^{}\mathbf{a}_{i,j}^l\mathbf{h}_j^l\mathbf{W}^l 
\end{equation}

\begin{equation} 
e\left(\mathbf{h}_{i},\mathbf{h}_{j}\right) = \textrm{LeakyReLU} \left(\mathbf{a}^{T}  
\cdot \left[\mathbf{W}\mathbf{h}_{i}\parallel\mathbf{W}\mathbf{h}_{j}\right]\right) 
\label{eq-GAT}
\end{equation} 

where attention scores $\mathbf{a}\in R^{2d^{'}}$ and weights $\mathbf{W}\in R^{d^{'}\times d}$ are learned. $\parallel$ denotes vector concatenation. Attention scores are normalized across all neighbours $ j\in\mathcal{N}_i$ using softmax, and the attention function is defined as:
\begin{equation} 
\begin{split}
\alpha_{ij} & = softmax_{j}\left(e\left(\mathbf{h}_{i},\mathbf{h}_{j}\right)\right) \\
& =\frac{exp\left(e\left(\mathbf{h}_{i},\mathbf{h}_{j}\right)\right)}{\sum_{j^{'}\in\mathcal{N}_{i}}^{} exp\left(e\left(\mathbf{h}_{i},\mathbf{h}_{j^{'}}\right)\right)} 
\end{split}
\end{equation} 

Finally, GAT computes a weighted average of the transformed features of the neighbour nodes (followed by a nonlinearity $ \sigma$) as the new representation of $i$, using the normalized attention coefficients:
\begin{equation} 
\mathbf{h}_{i}^{'} = \sigma\left(\sum_{j\in\mathcal{N}_{i}}^{}\alpha_{ij}\mathbf{W}\mathbf{h}_{j}\right) 
\end{equation} 

The motivation of GAT is to compute a representation for every node as a weighted average of its neighbours, by attending to its neighbours using its own representation as the query \cite{velivckovic2018graph}. Ability to focus on the most relevant parts of the input to make decisions results in robustness in the presence of noisy irrelevant neighbours \cite{alon2020bottleneck}.

\subsection{Dynamic attention}
\citet{brody2021attentive} notes that the main problem in the standard GAT scoring function, Equation~\ref{eq-GAT}, is that the learned layers $\mathbf{W}$ and $\mathbf{a}$ are applied consecutively, and thus can be collapsed into a single linear layer.

GATv2 replaces the linear approximator with a universal approximator function. 
\begin{equation} e\left(\mathbf{h}_{i},\mathbf{h}_{j}\right) =\mathbf{a}^T \cdot \textrm{LeakyReLU}\left(\mathbf{W}\left[\mathbf{h}_{i}\parallel\mathbf{h}_{j}\right]\right) 
\label{eq-GATv2}
\end{equation} 
Thus, GATv2 has been shown to perform better on noisy data \cite{brody2021attentive}.

\begin{figure*}[t]
  \includegraphics[width=\textwidth]{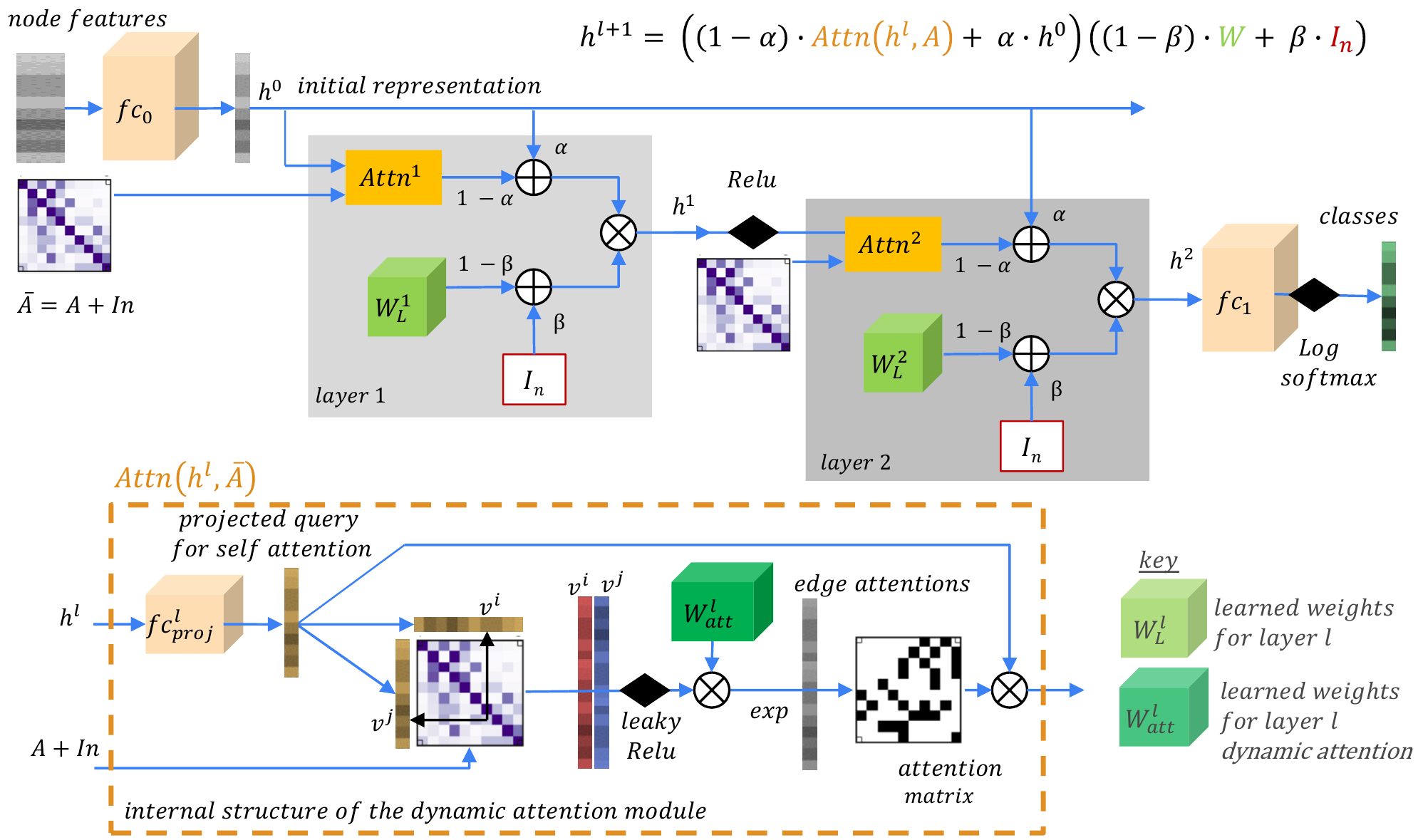}
  \caption{FDGATII uses an initial representation obtained from the node features via $fc_{0}$ projection combined with attention and Identity $I_{n}$ with $\alpha$ and $\beta$ proportions respectively at each layer. The attention module concatenates source (row) and destination (column) features of each edge from the adjacency matrix $A$ (with self-edges) projected via $W_{H}^{n}$, applies a non-linearity (leaky-relu) followed by an $exp()$ to obtain the edgewise attentions which is reshaped to a matrix suitable for the final softmax with the query. FDGATII can have multiple layers, followed by a final $W_{1}$ projection (layer) and log softmax that provides the node classification.}
  \label{fig_FDGATII}
\end{figure*}

\section{Our Proposed Architecture}
Most GNN models use a simple graph convolution based aggregation scheme as the basic building block \cite{kipf2016semi, hamilton2017inductive}. Recent studies point out that this leads to filter incompleteness which can be solved by using a more complex graph kernel \cite{abu2019mixhop}. Further, majority of the GNN models are designed under the assumption of homophily, and cannot handle heterophily \cite{zhu2020beyond}. Despite the GATs effectiveness, GAT performs poorly given heterophilic data, despite the attention mechanism’s inherent ability to focus on the most relevant nodes during the local aggregation process. 

Motivated by this limitations, we propose a modified attention based model that adapts well to heterohilic data while also solving oversmoothing in an efficient manner. 

Our proposed design (Figure~\ref{fig_FDGATII}) is built upon a local embedding step that extracts local node embeddings from the node feature vectors using GATv2. However, to extend the attention model to handle heterophilic and noisy data we borrow two techniques from GCNII \cite{chen2020simple} and H2GCN \cite{zhu2020beyond} with modifications, namely residual connections and identity mapping. We show, from a wide range of datasets, that this simple arrangement results in an efficient and fast model that generalizes well to homophilic and heterophilic datasets. 

Essentially, we combine GATv2 (Equation~\ref{eq-GATv2}) with Initial residual connection and Identity mapping as in Equation~\ref{eq-GCNII} to enhance local aggregation while ensuring robustness to heterophily. In Equation~\ref{eq-GCNII}, $\alpha$ and $\beta$ are the weights of Initial residual and the Identity respectively.

In addition to Equation~\ref{eq-GCNII}, GCNII also uses a variant, GCNII* with different weight matrices for the smoothed representation $\bar{P}\mathbf{H}^l$ and the initial residual $\mathbf{H}^{0}$. Formally, the $(l+1)$-th layer of GCNII* is defined as : 

\begin{equation}
\begin{split}
\mathbf{H}^{l+1} = \sigma\Bigg(\Bigg. \left(1-\alpha_l\right)\mathbf{\bar{P}H}^l\left(\left(1-\beta_l\right)\mathbf{I}_{n}+\beta_l\mathbf{W}_{1}^l\right) \\ 
+ \alpha_l\mathbf{H}^{0}\left(\left(1-\beta_l\right)\mathbf{I}_{n}+\beta_l\mathbf{W}_{2}^l\right) \Bigg.\Bigg)
\end{split}
\label{eq-GCNII2}
\end{equation}

We use both forms, ( Equation~\ref{eq-GCNII}, Equation~\ref{eq-GCNII2} ) of  supplement methods in our model.

\citet{chen2020simple} uses $\beta_l$ is to ensure the decay of the weight matrix adaptively increases with more layers. While our model, FDGATII is essentially a shallow model (typically 2 to 4 layers), we adopt the same settings of $\beta_l = log\left(\frac{\lambda }{l}+1\right)\approx\frac{\lambda }{l}$, where $\lambda$ is a hyperparameter. Further, we do not fine tune $\lambda$ for any dataset during model comparisons. \citet{zhu2020beyond} observations that (a) higher-order neighbourhoods; and (b) combination of intermediate representations improve the performance of models in heterophily settings and following MixHop \cite{abu2019mixhop}, which explicitly models 1-hop and 2-hop neighbourhoods, we use 2 to 4 layers. Following \citet{xu2018representation} we add skip connections in the form of initial representations $H^0$ as in \citet{chen2020simple}.

FDGATII differs from existing models with respect to its use of modified attention mechanism. Notably FDGATII is able to augment self attention with the use of a proportion of initial representation and identity resulting in an efficient shallow yet well genializing architecture. 

\section{Experiments}
In this section, we evaluate the performance of FDGATII against the state-of-the-art graph neural network models on a wide variety of open graph datasets for fully supervised classification.

Following \citet{pei2019geom} and \citet{chen2020simple} , we use 7 datasets (Table~\ref{table_dataset}). Cora, Citeseer, and Pubmed are homophilic citation network datasets where nodes correspond to documents, and edges correspond to citations; each node feature corresponds to the bag-of-words representation of the document and belongs to one of the academic topics. The remaining four (in Table~\ref{table_dataset}) are heterophilic datasets of web networks, where nodes and edges represent web pages and hyperlinks, respectively. The feature vector of each node is the bag-of-words representation of the corresponding page. 

Identical to \citet{pei2019geom} and \citet{chen2020simple} we use the same data splits, same mix of 60\%, 20\%, and 20\% for training, validation and testing, same pre-processing and measure the average model performance on the 10 test splits for each dataset. We fix the learning rate to 0.01, dropout rate to 0.5 and the number of hidden units to 64 on all datasets. Addressing observations by \citet{alon2020bottleneck} we do not perform  hyper-parameter fine tuning and use the benchmark, i.e.: GCNII, settings for comparability. For efficient computation, adjacency matrices are stored and used as sparse matrices.

We choose GCNII \cite{chen2020simple} as our performance and accuracy benchmark as it is (a) more current; (b) the most similar to our work in the use of initial representation and identity; (c) it actively attempts to solve the over smoothing problem and (d) as it is the current state-of-the-art in Cora dataset, a prominent data set for GNN model comparison. We also compare FDGATII with H2GCN \cite{zhu2020beyond} which is the current state-of-the-art on Cornel,Texas and Wisconsin; hightly heterophelic datasets.

\begin{table}[h]
    \caption{Statistics of the node classification datasets. H\% indicates the hormophily percentage.}
    \begin{tabular}{lllrrr}
    \toprule
    Dataset   & H\% & Cls. & Nodes  & Edges   & Features \\ \midrule
    Cora      & 0.81       & 7       & 2,708  & 5,429   & 1,433    \\
    Citeseer  & 0.74       & 6       & 3,327  & 4,732   & 3,703    \\
    Pubmed    & 0.80       & 3       & 19,717 & 44,338  & 500      \\
    Chameleon & 0.23       & 4       & 2,277  & 36,101  & 2,325    \\
    Cornell   & 0.30       & 5       & 183    & 295     & 1,703    \\
    Texas     & 0.11       & 5       & 183    & 309     & 1,703    \\
    Wisconsin & 0.21       & 5       & 251    & 499     & 1,703    \\ \bottomrule
%    PPI       &            & 121     & 56,944 & 818,716 & 50       \\ 
    \end{tabular}
    \label{table_dataset}
\end{table}

We further carryout training and inference time measurements with GPU warmup and proper synchronization prior to measurements. Additionally, for inference, we take the average time for 1000 inferences to lower any possibility of errors  and to be more reflective of real-world use of models.

%----all the result tables----------

\begin{table*}[!htbp]
\caption{Mean classification accuracy of full-supervised node classification. (1): Metrics reported by \citet{chen2020simple}, (2): Metrics reported by \citet{zhu2020beyond}, (3): GCNII best results from our tests with same pre-processing and 10 splits averaged as in \citet{chen2020simple}, (4): our FDGATII, with same pre-processing and 10 splits averaged as in \citet{chen2020simple} for comparison. (5) Accuracy  gain = $(GCNII_{acc} – FDGATII_{acc})*100 / GCNII_{acc}$. 
SOTA performance for each dataset is marked as bold and second-best performance is underlined for comparison. Parenthesis indicates the number of layers.}
\centering{
\begin{tabular}{llllllll}
\toprule
Method             & Cora               & Cite.     & Pumb.     & Cham.            & Corn.            & Texa.          & Wisc.          \\ \midrule
GCN\textsuperscript{1}               & 85.77              & 73.68     & 88.13     & 28.18            & 52.70            & 52.16          & 45.88          \\
GAT\textsuperscript{1}               & 86.37              & 74.32     & 87.62     & 42.93            & 54.32            & 58.38          & 49.41          \\
Geom-GCN-I\textsuperscript{1}        & 85.19              & 77.99     & 90.05     & 60.31            & 56.76            & 57.58          & 58.24          \\
Geom-GCN-P\textsuperscript{1}        & 84.93              & 75.14     & 88.09     & 60.90            & 60.81            & 67.57          & 64.12          \\
Geom-GCN-S\textsuperscript{1}        & 85.27              & 74.71     & 84.75     & 59.96            & 55.68            & 59.73          & 56.67          \\
APPNP\textsuperscript{1}             & 87.87              & 76.53     & 89.40     & 54.3             & 73.51            & 65.41          & 69.02          \\
JKNet\textsuperscript{1}             & 85.25(16)          & 75.85(8)  & 88.94(64) & 60.07(32)        & 57.30(4)         & 56.49(32)      & 48.82(8)       \\
JKNet(Drop)\textsuperscript{1}      & 87.46(16)          & 75.96(8)  & 89.45(64) & 62.08(32)        & 61.08(4)         & 57.30(32)      & 50.59(8)       \\
Incep(Drop)\textsuperscript{1}      & 86.86(8)           & 76.83(8)  & 89.18(4)  & 61.71(8)         & 61.62(16)        & 57.84(8)       & 50.20(8)       \\
GraphSAGE\textsuperscript{2}         & 86.90              & 76.04     & 88.45     & 58.73            & 81.18            & 82.43          & 75.95          \\
MixHop\textsuperscript{2}            & 87.61              & 76.26     & 85.31     & 60.50            & 75.88            & 77.84          & 73.51          \\
H2GCN-1\textsuperscript{2}           & 86.92              & 77.07     & 89.40     & 57.11            & \underline{82.16}      & \textbf{84.86} & \textbf{86.67} \\
GCNII\textsuperscript{1}             & \textbf{88.49(64)} & 77.08(64) & 89.57(64) & 60.61(8)         & 74.86(16)        & 69.46(32)      & 74.12(16)      \\
GCNII*\textsuperscript{1}            & 88.01(64)          & 77.13(64) & 90.30(64) & \underline{62.48(8)}   & 76.49(16)        & 77.84(32)      & 81.57(16)      \\ \midrule
GCNII\textsuperscript{3}            & 88.2696            & 76.9325   & 90.3499   & 63.75            & 77.2973          & 78.3784        & 79.8039        \\ 
FDGATII\textsuperscript{4}           & \underline{87.7867}      & 75.6434   & 90.3524   & \textbf{65.1754} & \textbf{82.4324} & \underline{80.5405}  & \underline{86.2745}  \\ \midrule
Accuracy Gain \%\textsuperscript{5} & -0.55              & -1.68     & 0.00      & 2.24             & 6.64             & 2.76           & 8.11           \\ \bottomrule
\end{tabular}
\label{table_acc}
}
\end{table*}

\begin{table*} [!htb] %[ht]   !htbp
\caption{Training and inference time comparison. In case of variants, we use the lowest average time taken to run all 10 standard splits. Efficiency = $GCNII_{time}/FDGATII_{time}$. We have used our time results of GCNII to eliminate any hardware-based effects. All timing is from running on Wiener supercomputer with CPU :Intel Xeon Gold 6132 GPU :V100/cuda/11.3.0 Memory:32G DDR4 allocation.}
\centering{
\begin{tabular}{llllllll}
\toprule
Metric               & Cora  & Cite. & Pumb.  & Cham. & Corn. & Texa. & Wisc. \\ \midrule
Training Time        &       &       &        &       &       &       &       \\
FDGATII (ms)         & 7489  & 1121  & 54247  & 3456  & 2892  & 4779  & 1842  \\
GCNII (ms)           & 79264 & 20070 & 246477 & 8540  & 13933 & 40224 & 11843 \\
Training Efficiency  & 10.6x  & 17.9x  & 4.5x    & 2.5x   & 4.8x   & 8.4x   & 6.4x   \\ \midrule
Inference Time       &       &       &        &       &       &       &       \\
FDGATII (ms)         & 4.96  & 4.92  & 6.97   & 3.65  & 3.62  & 4.70  & 3.74  \\
GCNII (ms)           & 35.06 & 35.37 & 52.64  & 6.01  & 8.90  & 16.32 & 8.86  \\
Inference Efficiency & 7.07x  & 7.19x  & 7.56x   & 1.65x  & 2.46x  & 3.47x  & 2.37x  \\ \bottomrule
\end{tabular}
\label{table_perf}
}
\end{table*}

\begin{table*} [!htbp]    %[ht]
\caption{Ablation study with and without II (Initial residual and Identity mapping). * Indicates the variant defined in Equation~\ref{eq-GCNII} and ** is from Equation~\ref{eq-GCNII2}. Each dataset is compared with the same hyperparameter settings (dropout, learning rate, decay) as in \citet{chen2020simple}. $L1$ and $L2$ identifies 1 and 2 layers respectively.}
\centering{
\begin{tabular}{llllllll}
\toprule
Metric           & Cora  & Cite. & Pumb. & Cham. & Corn. & Texa. & Wisc. \\ \midrule
Without   II, L1 & 86.90 & 75.65 & 87.01 & \textbf{65.18} & 65.95 & 62.16 & 54.51 \\
Without   II, L2 & 86.74 & 74.45 & 86.19 & 49.78 & 58.92 & 57.30 & 51.76 \\
With   II*, L1   & 87.06 & 75.07 & 89.96 & 61.34 & 76.76 & 70.00 & 81.96 \\
With   II*, L2   & \textbf{87.79} & \textbf{75.30} & \textbf{90.35} & 57.13 & 79.19 & 79.73 & 83.53 \\
With   II**, L1  & 84.91 & 75.28 & 89.48 & 49.12 & 80.27 & 78.65 & 84.12 \\
With   II**, L2  & 86.52 & 75.14 & 90.12 & 57.83 & \textbf{80.81} & \textbf{80.54} & \textbf{84.90} \\ \bottomrule
\end{tabular}
\label{table_able}
}
\end{table*}

\section{Results and discussion}
\subsection{Full-Supervised Node Classification}
Table~\ref{table_acc} reports the mean classification accuracy of each model. We reuse the metrics already reported by \citet{chen2020simple} and \cite{zhu2020beyond}. We observe that FDGATII demonstrates state-of-the-art results on heterophilic datasets while still being competitive on the homophilic datasets. Further FDGATII exhibits significant accuracy increases over its attention based predecessor, GAT. This result suggests that dynamic attention with initial residuals and identity improves the predictive power whilst keeping the layer count (and hence the model parameters and computational requirements) low. 

Table~\ref{table_perf} shows the training and inference time comparisons for our FDGATII with the benchmark GCNII model. FDGATII demonstrates up to 18x faster training speeds and up to 7x faster inference speeds. Figure~\ref{fig_performance} plots accuracy vs efficiency of FDGATII against the benchmark for all datasets, clearly indicating its superior performance mix.

\subsection{Ablation Study}
In this section, along with Table~\ref{table_able}, we consider the effect of various proposed design strategies on the performance of the model. Our 1 or 2-layer models, without Initial residual and Identity mapping (II), is theoretically equivalent to GAT/GATv2. The ablation study indicates that the addition of the II mechanism results in significant improvements on the heterophilic dataset performance. This result suggests that both II and dynamic attention techniques are needed to solve the problem of over-smoothing and data heterophily.   

\subsection{Suspended animation and over smoothing}
Table~\ref{table_smoothing} shows FDGATII performance for 2 selected datasets under increasing layer depth. As model depth increases, FDGATII does not exhibit performance degradation from suspended animation problem or over smoothing and performs well up to 32 layers of test depth.   

\begin{table}[h]
\caption{Model accuracy with increasing layer depth shows no presence of over smoothing or suspended animation problem. We used a hidden dimension of 64 and Equation~\ref{eq-GCNII} variant for all tests.}
\centering{
\begin{tabular}{@{}cll@{}}
\toprule
Num of Layers & Wisc. & Cite. \\ \midrule
1      & 81.96 & 75.07 \\
2      & 83.53 & 74.34 \\
4      & 82.16 & 74.57 \\
8      & 82.16 & 74.68 \\
16     & 81.76 & 74.86 \\
32     & 82.55 & 74.82 \\ \bottomrule
\end{tabular}
\label{table_smoothing}
}
\end{table}

\begin{table}[h]
\caption{Model layers and variant of II used for provided results. All other parameters are identical to \citet{chen2020simple}.}
\centering{
\begin{tabular}{@{}llrr@{}}
\toprule
Dataset   & Variant     & Dimensions & Layers \\ \midrule
cora      & Equation~\ref{eq-GCNII}  & 64         & 2      \\
citeseer  & Equation~\ref{eq-GCNII2} & 128        & 1      \\
pubmed    & Equation~\ref{eq-GCNII}  & 64         & 2      \\
chameleon & None                     & 64         & 1      \\
cornell   & Equation~\ref{eq-GCNII2} & 128        & 1      \\
texas     & Equation~\ref{eq-GCNII2} & 64         & 2      \\
wisconsin & Equation~\ref{eq-GCNII2} & 128        & 1      \\ \bottomrule
\end{tabular}
\label{table_parm}
}
\end{table}

\subsection{Performance and Efficiency}
\begin{figure}[h]
    \centering \includegraphics[width=1.00\columnwidth]{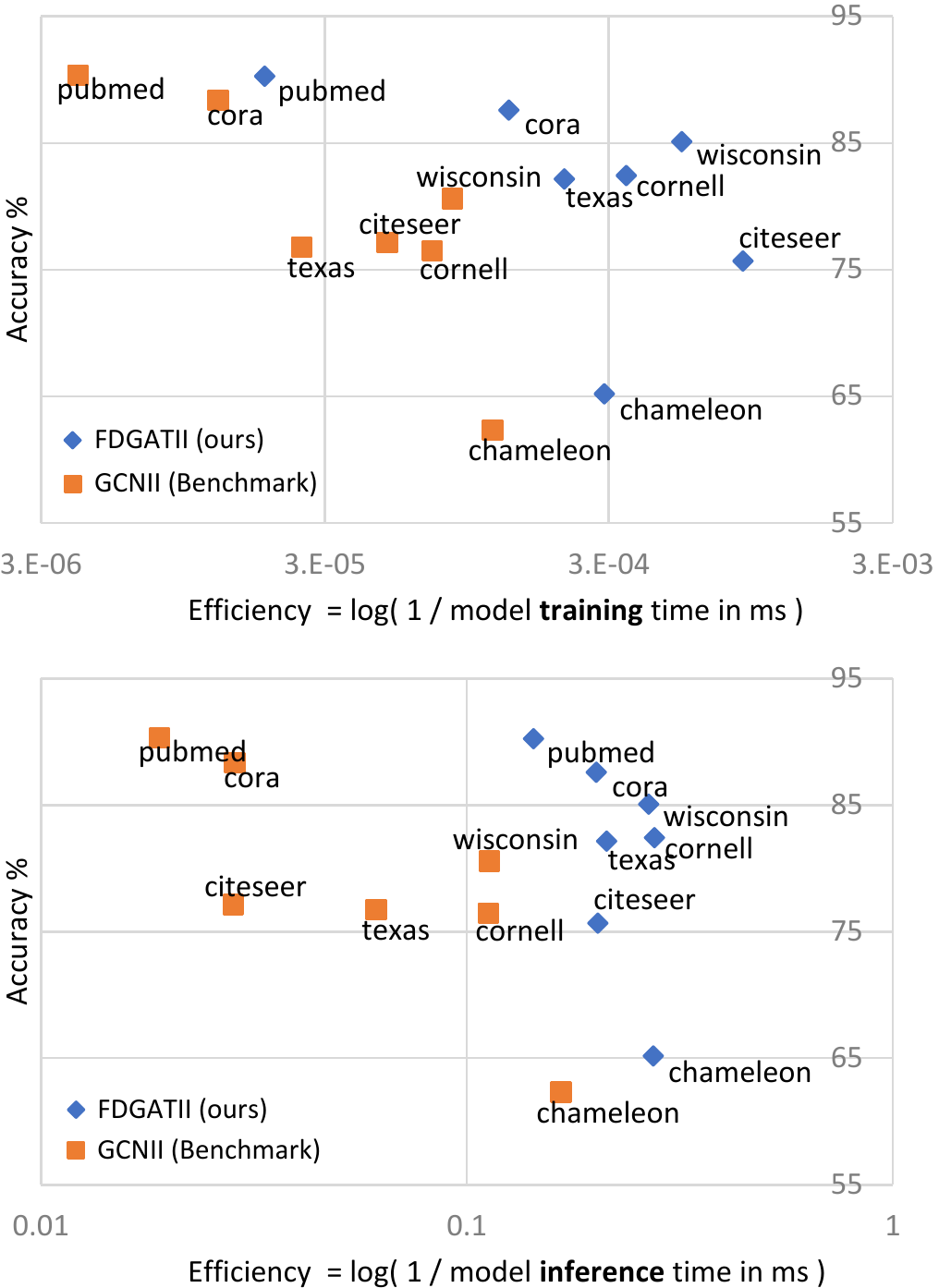}
    %\caption{Accuracy vs Efficiency. Area of the circle is proportionate to the inference time. Smaller circles indicate smaller inference times, i.e.: better inference efficiency}
    \caption{Training efficiency vs Accuracy (top); and Inference efficiency vs Accuracy (bottom). $Efficiency = log(1/ time)$ where training time is GPU time for 10 samples averaged with warm-up with GPU warm-up. Inference time is average for 1000 inferences on the GPU with warm-up.}
    \label{fig_performance}
\end{figure}

Figure~\ref{fig_performance} summarises the high accuracy to computational time efficiency ratio of FDGATII. For every dataset, FDGATT exhibits higher accuracy and/or lower training and inference times.  The proposed architecture performs consistently better across noisy and diverse datasets with comparable or better accuracy results to state-of-theart (Table~\ref{table_acc}) while exhibiting superiority in training and inference times, specifically 18x faster training speeds and up to 7x faster inference speeds over our chosen state-of-the-art benchmarks, GCNII \cite{chen2020simple} and H2GCN \cite{zhu2020beyond}. FDGATII's dynamic attention is able to achieve higher expressive power using less layers and parameters while still paying selective attention to important nodes, while the II mechanism supplements self-node features in highly heterophilic datasets. 

\subsection{Broader issues related to hetophily}
Many popular GNN models implicitly assume homophily producing results that may be biased, unfair or erroneous \cite{zhu2020beyond}. This can result in the so-called ‘filter bubble’ phenomenon  in a recommendation system (reinforcing existing beliefs/views, and downplaying the opposite ones), or make minority groups less visible in social networks creating ethical implications\cite{chitra2020analyzing}. FDGATI’s novel self-attention mechanism, where dynamic attention is supplemented with II for self-node feature preservation, reduces filter bubble phenomena and its potential negative consequences ensuring fairness and less bias.

\begin{figure}[h]
    \centering \includegraphics[width=0.8\columnwidth]{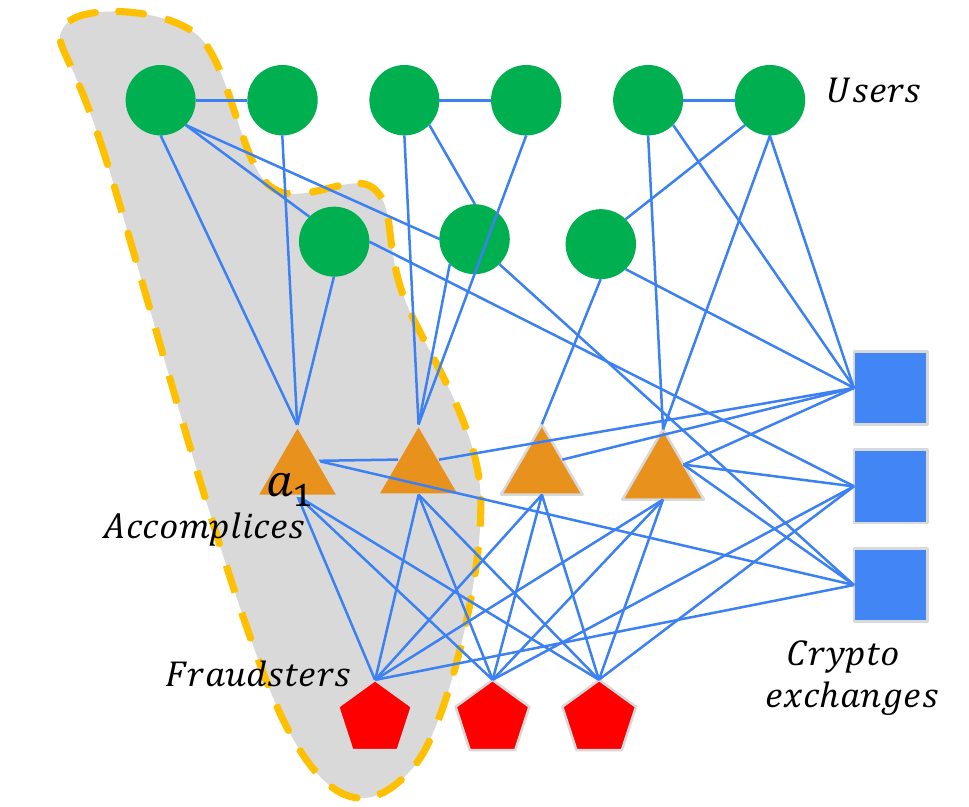} 
    \caption{Typical structure of a dark web market with illicit actors. Fraudsters would exhibit minimal links to fellow fraudsters and form form ties with other actors who play different roles \cite{bright2019illicit}. Simple aggregation of neighbours in such a scenario can be misleading \cite{alon2020bottleneck}.}
    \label{fig_hetro}
\end{figure}

This offers new possibilities for research into dataset where ‘opposites attract’ , i.e.: majority of linked nodes are different, such as social and dating networks (majority of gender connects with opposite gender), chemistry and biology (amino acids bond with dissimilar types in protein structures), e-commerce (sellers with promoters and influencers) and dark web and other cybercrime related activities \cite{zhu2020beyond}. In a typical dark web, fraudsters are more likely to connect to accomplices and prospective victims and not to fellow fraudsters, to facilitate efficiency while maintaining security, i.e.:  illicit actors will form ties with other actors who play different roles \cite{bright2019illicit}, resulting in heterophilic characteristics. Figure~\ref{fig_hetro} shows a typical set of nodes seen in a dark web marketplace with illicit actors.

Interesting future directions include combining FDGATII with fourier  based self-attention mechanisms \cite{lee2021fnet} and analysing the behaviour of FDGATII with novel attention mechanisms that integrate non-local information. 

\section{Conclusion}
This work investigates the use of modified self-attention models for full-supervised node classification. We propose FDGATII, a novel efficient  dynamic attention-based model that handles over-smoothing as well as robustness to noise (and heterophily) simultaneously by combining attentional aggregation with multiple feature preserving mechanisms based on initial residual connection and identity mapping. As a result, our model exceeds popular mainstream GAT and GCN benchmarks. Extensive experiments on wide spectrum of benchmark datasets shows that DFGATII achieves well-balanced state-of-the-art or second-best performance on various benchmark full-supervised tasks whilst exhibiting exceptional accuracy to efficiency ratios and parallelizability while, \textit{simultaneously} addressing over smoothing, suspended animation problem and heterophily prevalent in real world datasets.

% Acknowledgements should only appear in the accepted version.
\section*{Acknowledgements}

\bibliography{main}
\bibliographystyle{icml2021}

\end{document}